\documentclass[letterpaper, 10 pt, conference]{ieeeconf}  

\IEEEoverridecommandlockouts                              

\usepackage{graphics} 
\usepackage{epsfig} 
\usepackage{mathptmx} 
\usepackage{times} 
\usepackage{amsmath} 
\usepackage{amssymb}  
\usepackage[acronym]{glossaries}
\usepackage{caption}
\usepackage{subcaption}
\usepackage{stfloats}
\usepackage{xcolor}
\usepackage{cite}
\usepackage{mathtools}
\usepackage{hyperref}

\title{\LARGE \bf
Generating Stable and Collision-Free Policies through Lyapunov Function Learning
}

\author{Alexandre Coulombe$^{1}$ and Hsiu-Chin Lin$^{2}$
\thanks{$^{1}$Alexandre Coulombe is with the Department of Electrical and Computer Engineering, McGill University, Canada
        {\tt\small alexandre.coulombe@mail.mcgill.ca}}%
\thanks{$^{2}$Hsiu-Chin Lin is with the School of Computer Science and the Department of Electrical and Computer Engineering, McGill University, Canada
        {\tt\small hsiu-chin.lin@cs.mcgill.ca}}%
\thanks{This work is sponsored by Mitacs Accelerate IT23788. }
}

\usepackage{color}
\usepackage{amsmath}
\usepackage{amsfonts}
\usepackage{amssymb}
\usepackage[normalem]{ulem}

\newcommand{\dyn}[1]{\mathbf{\pi}\left(\pos{#1}\right)}
\newcommand{\posulim}{\mathbf{x}^+}
\newcommand{\posllim}{\mathbf{x}^-}
\newcommand{\velllim}{\dot{\mathbf{x}}^-}
\newcommand{\velulim}{\dot{\mathbf{x}}^+}
\newcommand{\start}{\mathbf{x}_0}
\newcommand{\lyapFunc}[1]{\tilde{V}\left({#1}\right)}

\newcommand{\R}{\mathbb{R}}
\newcommand{\dof}{d}

\newcommand{\ba}{\mathbf{a}}
\newcommand{\bb}{\mathbf{b}}
\newcommand{\bA}{\mathbf{A}}
\newcommand{\bB}{\mathbf{B}}

\newcommand{\pos}[1]{\mathbf{x}[#1]}
\newcommand{\velmax}{\dot{\mathbf{x}}_{max}}
\newcommand{\vel}[1]{\dot{\mathbf{x}}[#1]}
\newcommand{\goal}{\mathbf{x}_g}

\newcommand{\eigenNormal}[1]{\lambda_r\left(\pos{#1}\right)}
\newcommand{\eigenTangent}[1]{\lambda_e\left(\pos{#1}\right)}
\newcommand{\GammaFunc}[1]{\Gamma\left(\pos{#1}\right)}
\newcommand{\Mmat}[1]{\mathbf{M}\left(\pos{#1}\right)}
\newcommand{\Dmat}[1]{\mathbf{D}\left(\pos{#1}\right)}
\newcommand{\Emat}[1]{\mathbf{E}\left(\pos{#1}\right)}

\newacronym{STL}{STL}{Standard Triangle Language/Standard Tessellation Language}
\newacronym{URDF}{URDF}{Unified Robot Description Format}
\newacronym{AABB}{AABB}{Axis-Aligned Bounding Boxes}
\newacronym{OBB}{OBB}{Oriented Bounding Boxes}
\newacronym{GJK}{GJK}{Gilbert Johnson Keerthi}
\newacronym{RGJK}{RGJK}{Recursive Gilbert Johnson Keerthi}
\newacronym{EPA}{EPA}{Expanding Polytope Algorithm}
\newacronym{SVD}{SVD}{Singular Value Decomposition}
\newacronym{ANN}{ANN}{Artificial Neural Network}
\newacronym{RL}{RL}{Reinforcement Learning}
\newacronym{MDP}{MDP}{Markov Decision Process}
\newacronym{PPO}{PPO}{Proximal Policy Optimization}
\newacronym{IL}{IL}{Imitation Learning}
\newacronym{BC}{BC}{Behavioural Cloning}
\newacronym{DAGGER}{DAGGER}{Dataset Aggregation}
\newacronym{GAIL}{GAIL}{Generative Adversarial Imitation Learning}
\newacronym{IRL}{IRL}{Inverse Reinforcement Learning}
\newacronym{GCL}{GCL}{Guided Cost Learning}
\newacronym{GAN}{GAN}{Generative Adversarial Networks}
\newacronym{SEDS}{SEDS}{Stable Estimator of Dynamic Systems}
\newacronym{MSE}{MSE}{mean squared error}

\begin{document}

\maketitle
\thispagestyle{empty}
\pagestyle{empty}

\begin{abstract}
    The need for rapid and reliable robot deployment is on the rise. 
    \gls{IL} has become popular for producing motion planning policies from a set of demonstrations.
    However, many methods in \gls{IL} are not guaranteed to produce stable policies.
    The generated policy may not converge to the robot target, reducing reliability, and may collide with its environment, reducing the safety of the system. 
    \acrfull{SEDS} produces stable policies by constraining the Lyapunov stability criteria during learning, but the Lyapunov candidate function had to be manually selected.
    In this work, we propose a novel method for learning a Lyapunov function and a policy using a single neural network model. 
    The method can be equipped with an obstacle avoidance module for convex object pairs to guarantee no collisions. 
    We demonstrated our method is capable of finding policies in several simulation environments and transfer to a real-world scenario.
\end{abstract}
    
\begin{keywords}
Imitation Learning, Lyapunov stability, Obstacle Avoidance, Motion Planning, Neural networks
\end{keywords}

\section{INTRODUCTION}
\label{sec:Introduction}
\begin{figure}[t]
        \centering
        \begin{subfigure}[b]{0.3\linewidth}
            \centering
            \includegraphics[width=\linewidth]{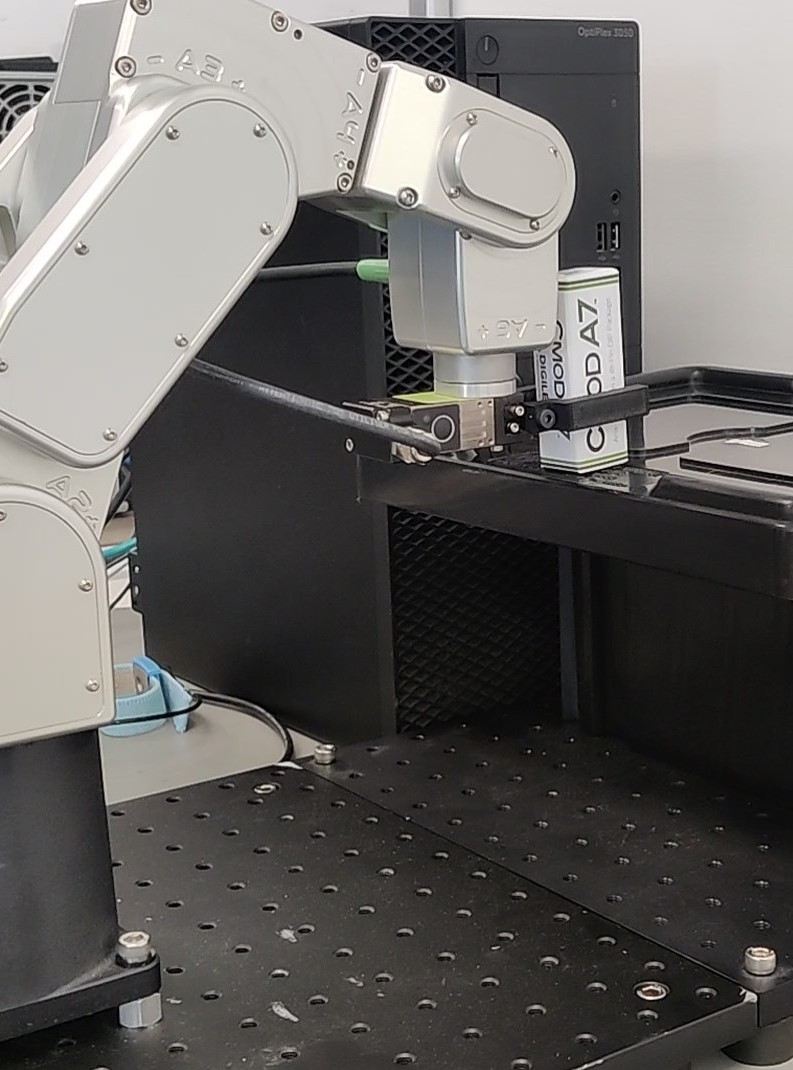}
            \caption{Initial Position}
            \label{fig:MecaInit}
        \end{subfigure}
        \begin{subfigure}[b]{0.3\linewidth}
            \centering
            \includegraphics[width=\linewidth]{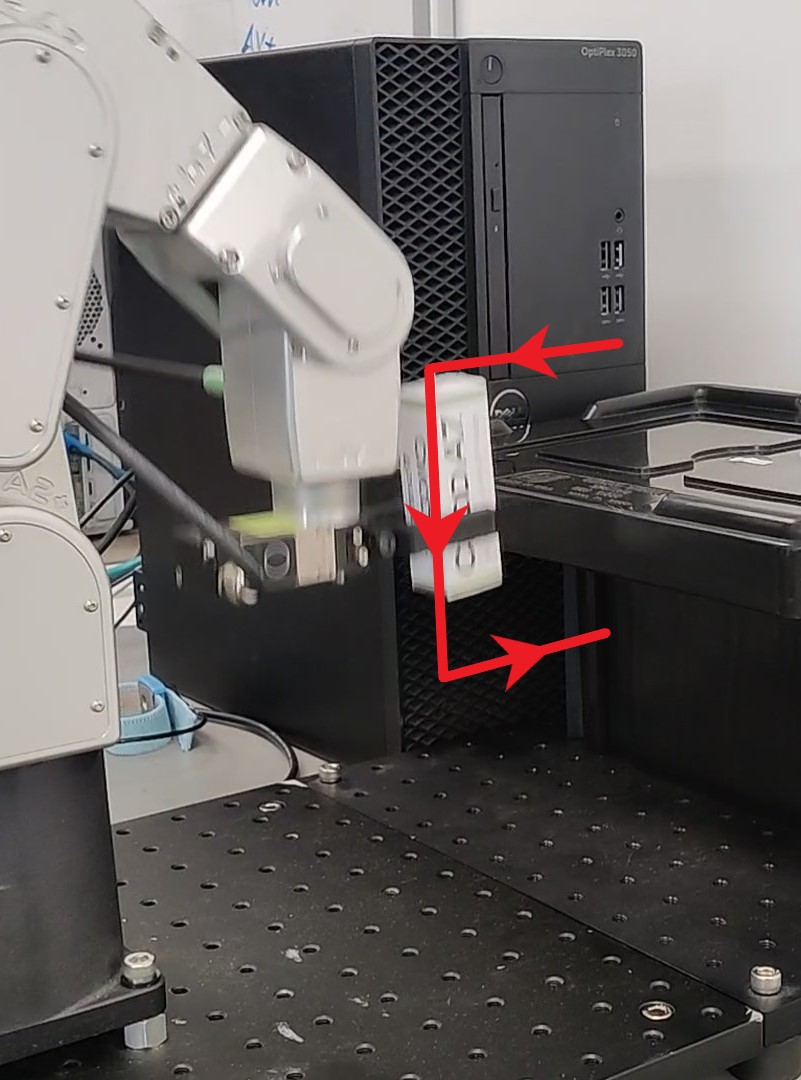}
            \caption{Policy Rollout}
            \label{fig:MecaTraj}
        \end{subfigure}
        \begin{subfigure}[b]{0.31\linewidth}
            \centering
            \includegraphics[width=\linewidth]{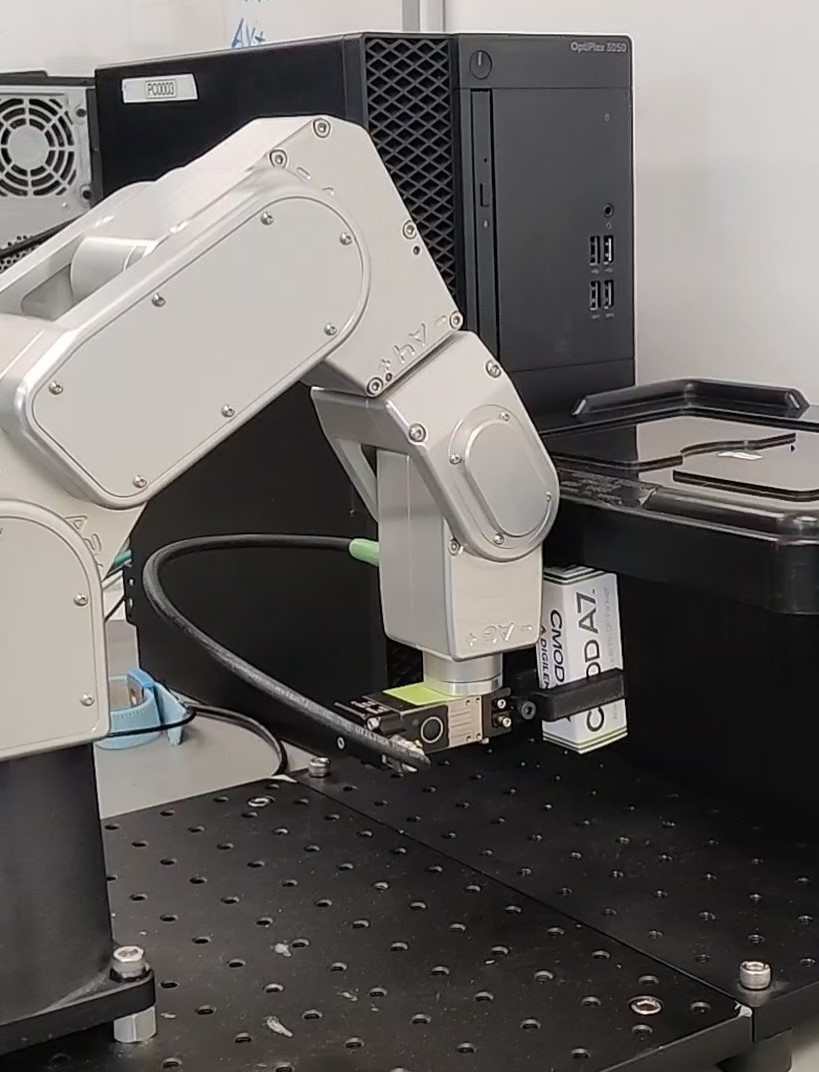}
            \caption{Goal Position}
            \label{fig:MecaGoal}
        \end{subfigure}
        \caption{Mecademic Meca500 performing the shelf manipulation task: the robot (a) starts from an initial position, (b) follows the policy rollout, and (c) reaches the goal position.}
        \label{fig:MecaEnv}
    \end{figure}

\noindent
    With the growing number of robots in industry and in human environments, the need for ease of deployment and safety in robotics becomes increasingly apparent. 
    Robots are skilled in routine and repetitive tasks. However, it is hard to handle a new task without tedious modelling, designing, and programming. 
    Furthermore, safety is crucial to deploying robotic systems since a mistake may injure the human or damage the robot itself. 
    This is particularly important when robots need to maneuver in environments with obstacles, such as the example in Figure~\ref{fig:MecaEnv}. 

    Trajectory optimization is a well-studied method for motion planning.
    The method finds a trajectory by minimizing some objective functions while satisfying its physical limits~\cite{TO.1998,STOMP.2011}.
    In the presence of obstacles, collision avoidance can be achieved by imposing a minimum distance between the robot and the obstacles as inequality constraints~\cite{schulman2013finding,schulman2014motion}. 
    While trajectory optimization is popular for motion planning, the solution is only valid for a specific pair of initial and target states. 
    Re-planning is a time-consuming process; therefore, it is not suitable for real-time planning. 
    Additionally, the planned path is not immune to perturbations.

    In recent years, \acrfull{IL}, the data-driven approach for motion planning, has shown promising results. 
    IL learns a model to predict the desired action given the current input, and therefore, produces a global solution as opposed to trajectory optimization.
    Many machine learning approaches have been applied, including supervised learning~\cite{bojarski2016end,lin2014novel},~\gls{IRL}~\cite{abbeel2004apprenticeship}, and \gls{GAIL}~\cite{ho2016generative, goodfellow2014generative}.
    However, these methods have no guarantees that the policy will be stable when deployed.

    Stability is crucial for robotic systems.
    Prior work proposed a data-driven approach with stability constraints to generate stable policies~\cite{khansari2011learning,figueroa2022locally}. 
    However,~\cite{khansari2011learning,figueroa2022locally} relies on a chosen Lyapunov candidate function to be valid for the demonstrations.
    Another prior work constrains the Lyapunov stability conditions on neural networks and learns the Lyapunov function with the policy from training data using mixed-integer programming~\cite{dai2021lyapunov}. 
    On the \gls{RL} front, 
    prior work introduces the stability constraint during training by widening the region of attraction of a Lyapunov function \cite{berkenkamp2017safe}. 
    However, the challenge is finding a suitable Lyapunov function. 

    In the presence of obstacles, methods such as virtual potential fields can modulate the policy to repel the robot from obstacles~\cite{warren1989global}. However, these create local minima where the robot is stuck.
    Recent work avoids extrema by modulating space around obstacles so that the robot moves around them, but it only works for point representations of the robot~\cite{khansari2012dynamical, huber2019avoidance, huber2022avoiding}.

    In this paper, we propose a novel method for learning both the Lyapunov function and policy with a single neural network.
    To produce a stable policy, we train the network using a constrained optimization formulation with stability conditions as constraints.
    We deploy the trained policy with an obstacle avoidance module, augmented by our method to deal with convex objects. 
    The proposed work is validated in simulation and robotic hardware with direct sim-to-real transfer. 
    The main contributions include:
    \begin{itemize}
        \item We extend~\cite{khansari2011learning} by learning a Lyapunov function from demonstrations instead of manually choosing a Lyapunov candidate function (Sec.~\ref{sec:Method}).
        \item We extend the obstacle avoidance work in~\cite{khansari2012dynamical, huber2019avoidance, huber2022avoiding} to treat robots and obstacles as {\em convex objects} (Sec.~\ref{sec:ObsModule}).
        \item We show that policies developed in simulation can be transferred to a real-world system reliably (Sec.~\ref{sec:Experiment}).
    \end{itemize}
    
\section{BACKGROUND}
\label{sec:Background}
\noindent 
In this section, background knowledge for Lyapunov stability theory and collision avoidance techniques are provided.

\subsection{Lyapunov Function}
\label{sec:LyapunovFunction}
\noindent
    From~\cite{stability}, Lyapunov stability dictates that a system with $\dof$ degree-of-freedom is locally stable if, for a region around the equilibrium $\goal\in\R^\dof$, there is a Lyapunov function $V(\cdot): \R^\dof \rightarrow \R$ that satisfies the conditions:
    \begin{align}
        &\label{eq:LyapunovPositive} V(\pos{n}) > 0, &\ \forall \pos{n} \in X, \pos{n} \neq \goal\\
        \label{eq:LyapunovDec}  
        &V(\pos{n+1}) \leq (1 -\epsilon) V(\pos{n}), &\ \forall \pos{n} \in X, \pos{n} \neq \goal\\
        \label{eq:LyapunovGoal} 
        & V(\goal) = 0 &
    \end{align}
    where $\epsilon > 0$ is a positive scalar and $X = \{ \pos{0} \; | \; V(\pos{0}) \leq \rho \}$ is the stable region around $\goal$, where $\rho > 0$ is some positive value. 
    \eqref{eq:LyapunovPositive} and~\eqref{eq:LyapunovGoal} specify that $V(\cdot)$ must output a positive value for all positions, except for $0$ at $\goal$.
    \eqref{eq:LyapunovDec} states that the system must evolve to positions with lower values, with the lowest value at $\goal$. 
    Thus, the system will converge to the equilibrium point $\goal$ if a Lyapunov function can be found.
    
\begin{figure}[t]
        \centering
        \begin{subfigure}[t]{0.475\linewidth}
            \centering
            \includegraphics[height=2.8cm]{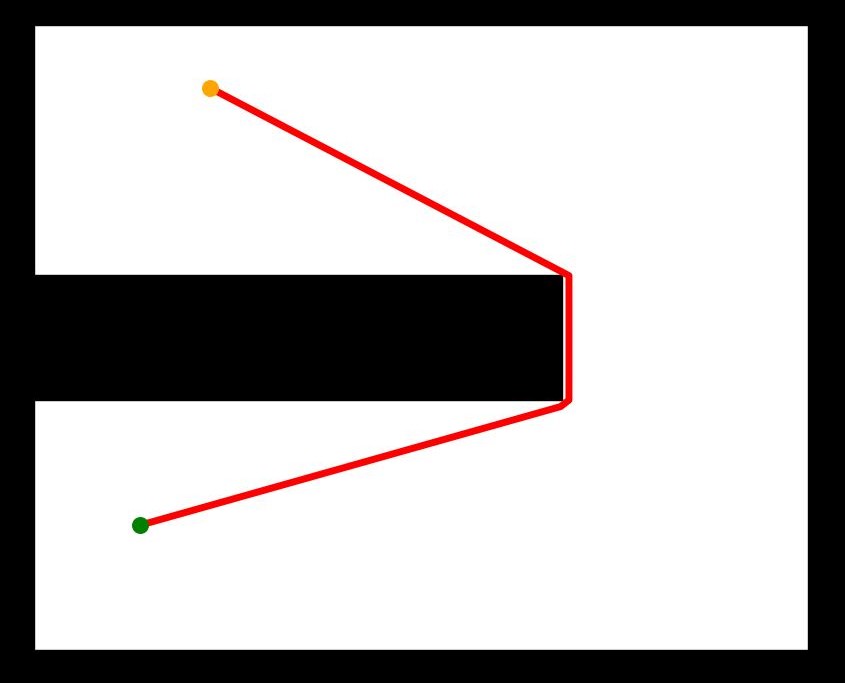}
            \caption{Demonstrations with an initial (orange) and a goal (green) states while avoiding obstacles (black).}    
        \end{subfigure}
        \hfill
        \begin{subfigure}[t]{0.475\linewidth}
            \centering
            \includegraphics[height=2.4cm]{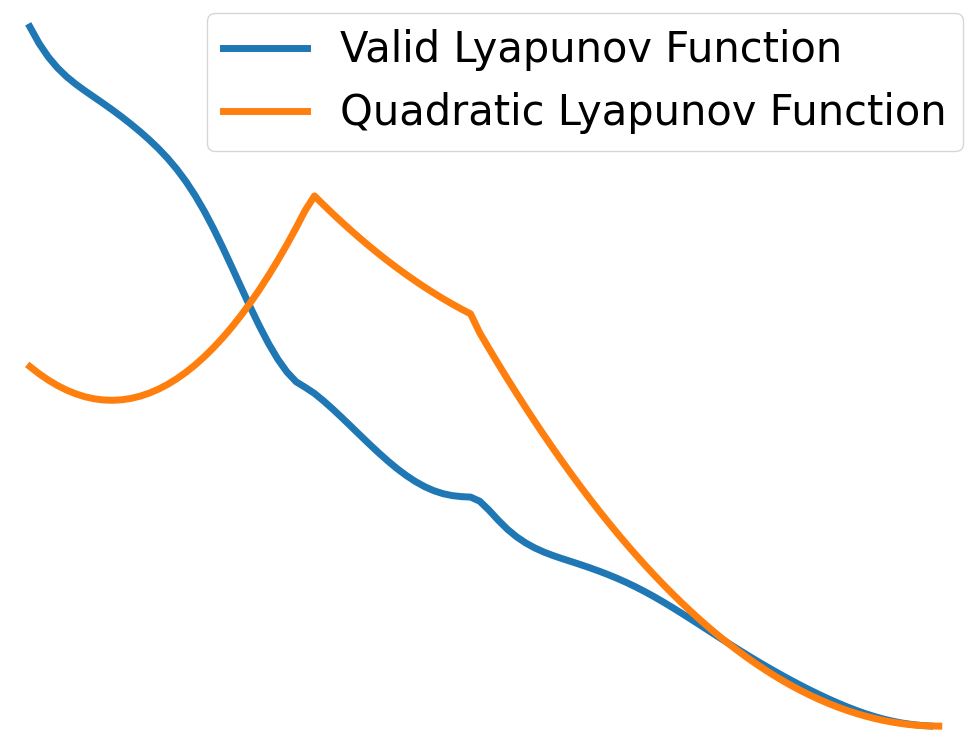}
            \caption{Outputs of a quadratic Lyapunov function (orange) and a valid Lyapunov function (blue) for the demonstrations.}    
        \end{subfigure}
        \caption{An example of a trajectory and its Lyapunov function outputs. 
        }
        \label{fig:ValidLyapunovFunc}
    \end{figure}

    A common Lyapunov candidate function is the quadratic Lyapunov function 
    \begin{equation}
         V(\pos{n}) = \| \pos{n} - \goal \|_2
         \label{eq:quadLyap}
    \end{equation}
    The above Lyapunov candidate function imposes constraints on the movement towards the goal. However, this may not be ideal in some scenarios, such as moving around obstacles. 
    An example is shown in Figure~\ref{fig:ValidLyapunovFunc}. 
    The quadratic Lyapunov candidate function (orange) violates the constraint in~\eqref{eq:LyapunovDec}, since avoiding the obstacles implies moving away from the goal, while a valid Lyapunov function (blue) is monotonically decreasing along the demonstration path (red).
    
    Instead of explicitly specifying the Lyapunov candidate function, prior work~\cite{dai2021lyapunov} learns a function that satisfies the Lyapunov conditions in~\eqref{eq:LyapunovPositive}-\eqref{eq:LyapunovGoal}. In this work, we will adapt this structure to learn the Lyapunov function. 

\subsection{Collision Avoidance}
\label{sec:CollisionAvoidance}

    \noindent
    Given the distance between the robot and the obstacle, a policy $\dyn{n}$ can be modulated to move around the obstacles~\cite{khansari2012dynamical, huber2019avoidance, huber2022avoiding} with the following formulation:
    \begin{equation}
    \begin{aligned}
        \vel{n} &= \Mmat{n}\dyn{n}
        \label{eq:ModulationMat}
    \end{aligned}
    \end{equation}
    where $\Mmat{n}$ is the modulation matrix of the obstacle. 
    The modulation matrix conserves the existing extrema of the system dynamics, and does not introduce new extrema while the matrix is full rank \cite{huber2019avoidance}. 
    The modulation matrix $\Mmat{n} $ is formed by the basis matrix $\Emat{n} $ and eigenvalue diagonal matrix $ \Dmat{n} $
    \begin{equation}
        \begin{aligned}
            \Mmat{n} &= \Emat{n}\Dmat{n}\Emat{n}^{-1} \\
            \Emat{n} &= \left[r\left(\pos{n}\right), e_1\left(\pos{n}\right), \dots,  e_{d-1}\left(\pos{n}\right)\right]\\
            \Dmat{n} &= \mathbf{diag}\left(\eigenNormal{n}, \eigenTangent{n}, \dots,  \eigenTangent{n}\right)
        \end{aligned}
        \label{eq:modulation-matrix}
    \end{equation}
    
    \noindent where 
    $r\left(\pos{n}\right)$ is a vector from the obstacle reference point to the robot reference point, and
    $e_i \left( \pos{n} \right)$ are vectors tangent to the obstacle surface.
    $\eigenNormal{n}, \eigenTangent{n}$ are related to the distance between the robot and the obstacle $\GammaFunc{n}$ as: 
    \begin{equation}
        \eigenNormal{n} = 1 - \frac{1}{\GammaFunc{n}},\quad \eigenTangent{n} = 1 + \frac{1}{\GammaFunc{n}}
        \label{eq:eigenvalues}
    \end{equation}
    
    \noindent
    This method prevents motion from penetrating the surface while encouraging the system to move in the tangent directions of the obstacles. 
    In this work, we extend this method by treating convex obstacles with a convex hull representation of the robot.
    
\subsection{Distance Measurement for Convex Objects}
\noindent
The collision avoidance technique in the previous section requires the signed distance between the robot and an obstacle.  
Assuming the robot and the obstacle can be described as a convex object, the distance between two convex objects $\bA$ and $\bB$ can be calculated with the Minkowski difference,
$\bA \ominus \bB = \{ \ba-\bb : \forall \ba \in \bA , \forall \bb \in \bB \}$.

To find the shortest distance between $\bA$ and $\bB$ without computing the entire Minkowski difference,~\gls{RGJK}~\cite{gilbert1988GJK} and~\gls{EPA}~\cite{van2003collision} exploit its properties to obtain the shortest separating and penetrating distance respectively. 
Together, they give the signed distance $sd(\bA,\bB)$ between a pair of convex hulls as follows:
\begin{equation}
    sd(\bA,\bB) = RGJK(\bA, \bB) - EPA(\bA, \bB)
    \label{eq:convexHullSD}
\end{equation}
Many extensions were introduced to speed up this computation~\cite{ong1997GJK, coulombe2020collision}.

\section{PROPOSED APPROACH}
\label{sec:Approach}

\noindent
We consider a task with $\dof$ degree-of-freedom where $\pos{\cdot}\in\R^\dof$ represents the state, and $\vel{\cdot}\in\R^\dof$ represents the velocities. 
Assuming we have full knowledge of the environment, 
our goal is to learn a policy $\vel{\cdot}=\pi(\pos{\cdot}): \R^\dof \rightarrow \R^\dof$ that outputs the velocities which guides the robot towards a target state $\goal\in\R^{\dof}$ while enforcing stability on the motion.

\subsection{Automated Demonstration Collection}
\label{sec:Method-to}
\noindent
    The data is automatically generated via trajectory optimization.
    Given an initial position $\start$ and a target position $\goal$, the objective of the optimization is to find a trajectory of positions $\pos{\cdot}$ and a trajectory of velocities $\vel{\cdot}$, that travel from $\start$ and $\goal$ with the shortest distance.
    In addition, the solution should be constrained by the position limits $\posllim, \posulim$ and the velocity limits $\velllim, \velulim$.
    
    Trajectory optimization allows us to introduce additional constraints such as avoiding collisions with the environment.
    Assuming that the robot has $N_{link}$ links and performs some tasks in an environment of $N_{obstacles}$ obstacles, we need to avoid collisions between each link of the robot and the obstacles. 
    For collision avoidance, we model each object as a convex hull, since they are less prone to overestimating the volume of the object. For non-convex objects, the object can be represented as a collection of its convex components. 
    The trajectory optimization is formulated as follows: 
    \begin{equation}
    \begin{aligned}
        \min_{\pos{\cdot}, \vel{\cdot}} &~~~ \sum_{n=0}^{N-1} \| \pos{n+1} - \pos{n} \|  \\
        \text{subject to~}  & \pos{n+1} = \mathbf{f}(\pos{n}, \vel{n}), \forall n \in [0, N-1] \\
                            & \pos{0} = \start, \quad \pos{N} = \goal \\
                            & \posllim \leq \pos{n} \leq \posulim \\
                            & \velllim \leq \vel{n} \leq \velulim \\
                            & sd(A_i,O_j) \geq d_{safe}, \forall i \in [ 1, N_{link} ] \\
                            & \hspace{60pt} \forall j \in [1, N_{obstacles ]}
    \end{aligned}
    \label{eq:TrajOptDemo}
    \end{equation}
    where $N$ is the number of via points, $ \mathbf{f}$ is the forward dynamics of the robot, $sd$ is the signed distance for convex hulls in \eqref{eq:convexHullSD}, $A_i$ is the convex hull of the $i^{th}$ link of the robot, $O_i$ is the convex hull of the $i^{th}$ obstacle, and $d_{safe} > 0$ is the minimum distance that must not be violated for obstacles.
    
    With this information, the formulation in \eqref{eq:TrajOptDemo} can be passed to an interior point optimization solver, such as IPOPT~\cite{wachter2006implementation}. This can be run multiple times from different initial positions to produce a set of demonstrations.

\subsection{Value Function and Policy Learning}
\label{sec:Method}
\noindent
    We assume that data are generated using Sec.~\ref{sec:Method-to} as a set of positions $\pos{\cdot}$ and velocities $\vel{\cdot}$.
    Our goal is to learn a policy $\pi$ that predicts the most suitable velocity given the current state $\vel{\cdot}= \pi(\pos{\cdot})$ and
    satisfies the Lyapunov stability conditions. 
    The proposed supervised learning method aims to shape a Lyapunov function $\lyapFunc{\cdot}: \R^\dof \rightarrow \R$.
  
    We follow the formulation in~\cite{dai2021lyapunov} to structure a Lyapunov candidate function as follows:
    \begin{equation}
        \lyapFunc{\pos{n}} = \phi\left(\pos{n} - \goal, \theta_V \right) - \phi\left(\mathbf{0}, \theta_V\right) +    \|\pos{n} - \goal\|  
    \end{equation}
    where 
    $\phi(\cdot,\theta_V ): \R^\dof \rightarrow \R$ is an \gls{ANN} parameterized by $\theta_V$.
    This formulation imposes that the value at the target state is $0$, which satisfies \eqref{eq:LyapunovGoal}.

     \begin{figure}[t]
        \centering
        \begin{subfigure}[b]{0.625\linewidth}
            \centering
            \includegraphics[width=0.9\linewidth]{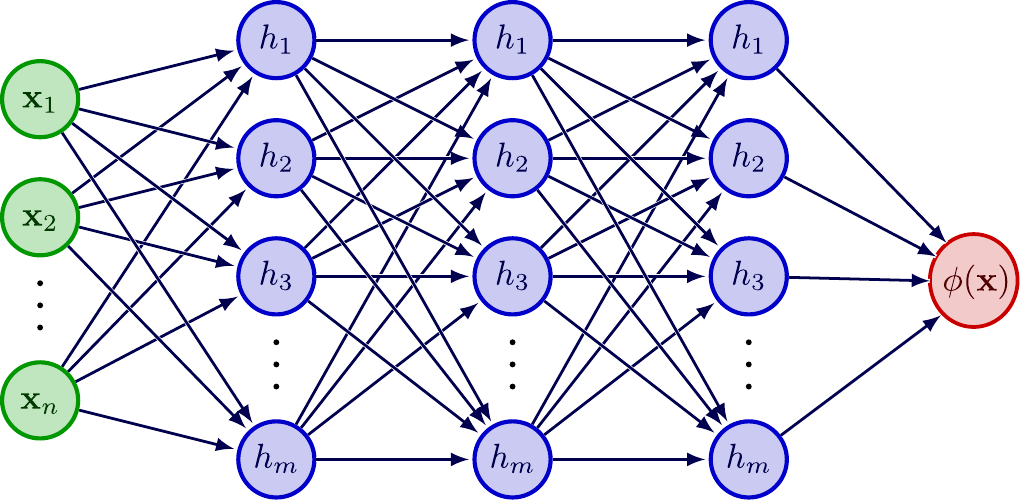}
            \caption{\gls{ANN} for the Lyapunov function}
            \label{fig:method-network}
        \end{subfigure}
        \begin{subfigure}[b]{0.325\linewidth}
            \centering
            \includegraphics[width=0.9\linewidth]{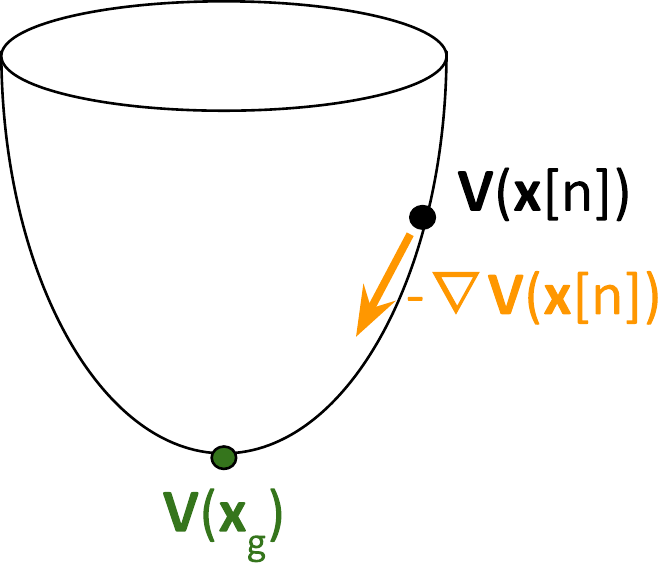}
            \caption{Policy}
            \label{fig:method-lyapunov}
        \end{subfigure}
        \caption{Learning Lyapunov policy: (a) an \gls{ANN} is learned (b) the action is the negative gradient of the Lyapunov function}
        \label{fig:LFL_method}
    \end{figure}
    
    The policy can be taken to be related to the negative gradient of the learned Lyapunov function
    \begin{equation}
        \pi\left(\pos{n}\right) \propto -\nabla \lyapFunc{\pos{n}}
    \end{equation}
    \noindent
    The idea is illustrated in Fig.~\ref{fig:LFL_method}. 
    The Lyapunov function $\lyapFunc{\pos{n}}$ is nonlinear with respect to positions $\pos{n}$, so we deploy an~\gls{ANN} to learn the relationship (Fig.~\ref{fig:method-network}). 
    Once a Lyapunov function is learned, the negative gradient $ -\nabla \lyapFunc{\pos{n}}$ will guide the robot toward the target (Fig.~\ref{fig:method-lyapunov}). 
    

    To train the neural network, the objective is to minimize the discrepancy between the policy output $-\nabla \lyapFunc{\pos{n}}$ and the demonstrated action $\vel{n}$.
    This discrepancy $e(n)$ can be taken from the dot product between two vectors. 
    \[
       -\nabla  \lyapFunc{\pos{n}} \cdot \vel{n} =  \|\nabla  \lyapFunc{\pos{n}}\|  \|\vel{n}\| \cos(e(n)) 
    \]
    
    \noindent 
    Dividing both sides by the magnitude of the two vectors and find the inverse cosine, the objective function becomes
    \begin{equation}
        \min_{\theta_V} \sum_{n=0}^{N-1} \Bigg\|\arccos{\left(\frac{-\nabla  \lyapFunc{\pos{n}} \cdot 
        \vel{n}}{\|\nabla  \lyapFunc{\pos{n}}\| \|\vel{n}\|}\right)}\Bigg\| 
        \label{eq:objective}
    \end{equation}
  
    \noindent
    In order to generate a valid Lyapunov function, the stability conditions in~\eqref{eq:LyapunovPositive} and~\eqref{eq:LyapunovDec} are added to the objective in~\eqref{eq:objective}.
    The problem is converted into:
    \begin{equation}
    \begin{split}
        \min_{\theta_V} &\sum_{n=0}^{N-1} \Bigg\|\arccos{\left(\frac{-\nabla  \lyapFunc{\pos{n}} \cdot \vel{n}}{\|\nabla  \lyapFunc{\pos{n}}\| \|\vel{n}\|}\right)}\Bigg\| \\
        \text{subject to}\ &  \lyapFunc{\pos{n}} > 0, \forall \pos{n} \neq \goal,\\
        &  \lyapFunc{\pos{n+1}} - (1 - \epsilon) \lyapFunc{\pos{n}} \leq 0
        \label{eq:VLfD_opt}
    \end{split}
    \end{equation}
    
    \noindent 
    To make the optimization easier for implementation in tools such as PyTorch \cite{Pytorch}, the formulation is modified to be an augmented Lagrangian. Also, the trigonometry is simplified to be a difference by rearranging the terms of the dot product and inserting the desired angle. These changes produce the following optimization:
    \begin{equation}
    \begin{aligned}
        \min_{\theta_V} & \sum_{n=0}^{N-1} \left( 1 + \frac{\nabla  \lyapFunc{\pos{n}} \cdot \vel{n}}{\|\nabla \lyapFunc{\pos{n}}\| \|\vel{n}\|}\right)\\
       & + \lambda_1 \max \Big( -  \lyapFunc{\pos{n}}, 0 \Big)\\
       & + \lambda_2 \max \Big( \lyapFunc{\pos{n+1}} - (1 - \epsilon)  \lyapFunc{\pos{n}}, 0 \Big) 
    \end{aligned}
    \label{eq:VLfD_train}
    \end{equation}
    where $\lambda_1$ and $\lambda_2$ are the Lagrange multipliers. 
    The model will be a valid Lyapunov function for the region around the provided demonstrations. 
    Outside the demonstration regions, there is no guarantee on the stability of the model. Termination before the constraints are satisfied will result in an invalid Lyapunov function. 
    Finally, to ensure that the policy output is within the physical limits, the output is scaled by
    \begin{equation}
        \pi\left(\pos{n}\right) = \frac{-\nabla \lyapFunc{\pos{n}}}{\|\nabla  \lyapFunc{\pos{n}}\|} \velmax
    \end{equation}

\subsection{Obstacle Avoidance for Convex object pairs}
\label{sec:ObsModule}
\noindent
    The learning approach introduced in the previous section learns a policy that reproduces the demonstrated data with stability guarantee. 
    However, it has no information about obstacles in the environment. 
    To further guarantee a collision-free policy, the obstacle avoidance method from~\cite{huber2022avoiding} was improved upon. 
    While~\cite{huber2022avoiding} treats the robot as a {\em point}, we augmented the method for handling {\em convex representations}.
    We represent each link of the robot with the convex hull of the link, and the obstacles are convex objects. 
    
    \begin{figure}[t]
        \centering
        \includegraphics[width=0.6\linewidth]{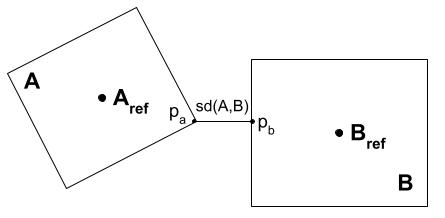}
        \caption{The reference points on the convex hulls used to get the distance function output.}
        \label{fig:convexhullDist}
    \end{figure}
    For convex to convex comparison, the signed distance between the convex objects is required. 
    An example is shown in Fig.~\ref{fig:convexhullDist}. Assuming $\bA$ and $\bB$ are two convex objects, $\bA_{ref}$ and $\bB_{ref}$ are the reference position of object $\bA$ and $\bB$ respectively, $p_a$ is the closest point on $\bA$ to $\bB$, and $p_b$ is the closest point on $\bB$ to $\bA$. 
    The signed distance $sd(\bA,\bB)$ between $\bA$ and $\bB$ can be computed from~\eqref{eq:convexHullSD}. 
    
    For handling convex objects, we define the distance function for the modulation matrix between convex objects as 
    \begin{equation}
        \GammaFunc{n} = \frac{\|\bA_{ref} - \bB_{ref}\|}{\|\bA_{ref} - \bB_{ref}\| - sd(\bA, \bB)}
        \label{eq:new-gamma}
    \end{equation}
    In~\eqref{eq:new-gamma}, the signed distance $sd(\bA,\bB)$ is used to form a ratio.
    When $sd(\bA, \bB)= 0$, the two objects collide. 
    The result is $\GammaFunc{n}=1$, so that $\eigenNormal{n} = 0$ in~\eqref{eq:eigenvalues} and makes~\eqref{eq:ModulationMat} block motion towards the obstacle.
    When $sd(\bA,\bB) \ge 0$, $\GammaFunc{n} \ge 1$, which gives the desired eigenvalues for the obstacle modulation when they are not intersecting. 
    
For the basis vectors in ~\eqref{eq:modulation-matrix}, $r\left(\pos{n}\right) = p_a - p_b / \|p_a - p_b\|$ is taken to be the normal direction in which to impede robot motion and $e_i \left( \pos{n} \right)$ are chosen to be orthogonal to the normal. In 2D, this can simply be the normal rotated by $90^{\circ}$. In 3D, this can be done using spherical coordinates and taking $\hat{r} = r\left(\pos{n}\right)$, and $\hat{\theta}$ and $\hat{\phi}$ to be the tangents.  

\begin{figure}[t]
        \centering
        \includegraphics[width=\linewidth]{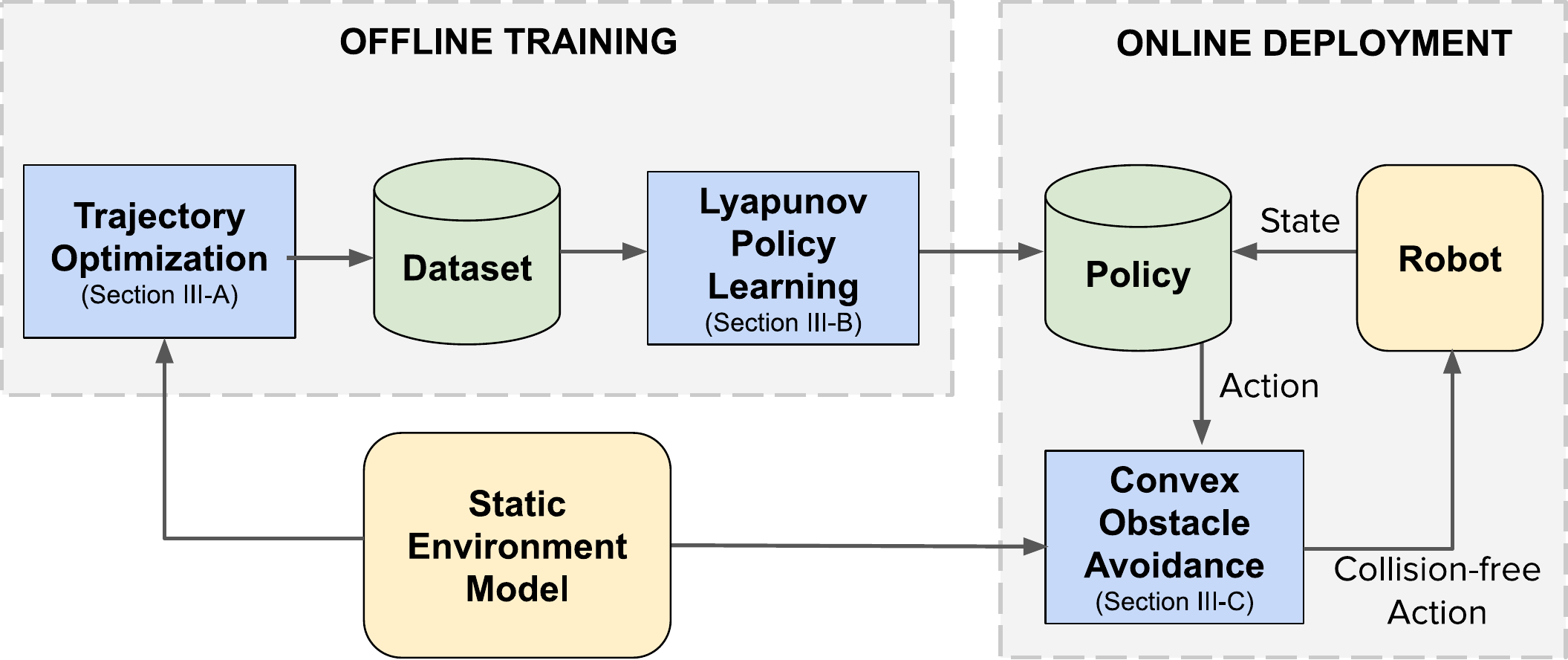}
        \caption{A summary of our proposed method}
        \label{fig:flowchart}
    \end{figure}

A summary of our proposed method is illustrated in Fig.~\ref{fig:flowchart}.

\section{EXPERIMENTS}
\label{sec:Experiment}
\noindent
    To verify the performance of the proposed method, we evaluate the method on several different environment scenarios. 
    The scenarios explored are 2D examples, a manipulation task in simulation and on a real robot. 
    For each case, a set of demonstrations are collected through trajectory optimization method from Section~\ref{sec:Method-to}. 
    We use LASA \gls{SEDS} as our baseline, where the chosen Lyapunov function is~\eqref{eq:quadLyap}. 
    Details about the baseline and our implementations can be found in the Appendix. We used the obstacle avoidance module in all experiments to show the difference in learned system dynamics. 
    
    \subsection{Evaluation Criteria}
    \noindent
    The performance of the method is evaluated based on its convergence to the goal, its ability to avoid collisions, its ability to satisfy the Lyapunov stability conditions and the prediction error.
    The prediction error is taken as the \gls{MSE} of the unit vectors of the policy outputs and demonstration actions.
    
\subsection{Experiment 1: 2D examples}
    \begin{figure*}[t]
        \centering
        \begin{subfigure}[b]{0.3\textwidth}
            \centering
            \includegraphics[width=\linewidth]{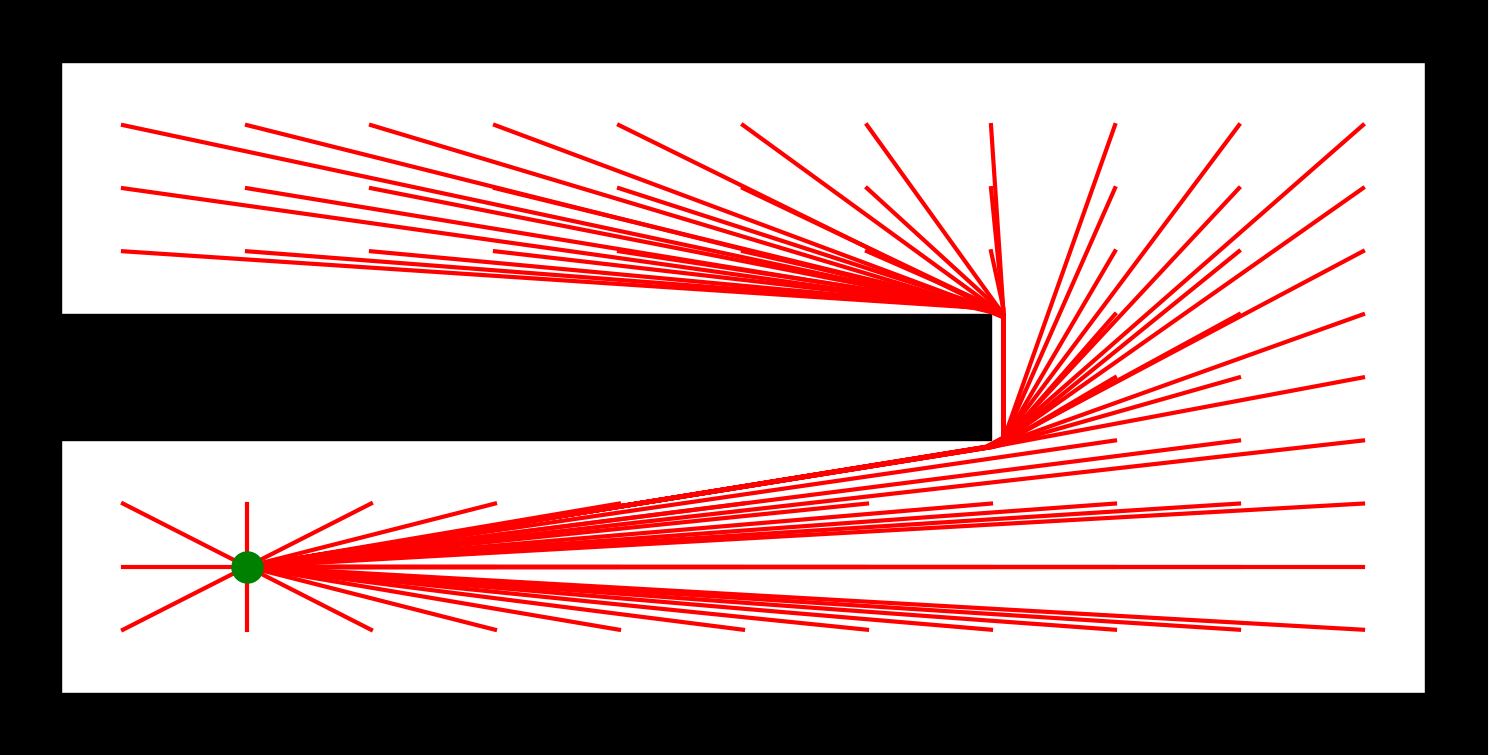}
            \caption{Demonstrations (red arrows) with the goal is marked with a green dot.}
            \label{fig:Hall_Demo}
        \end{subfigure}
        \hfill
        \begin{subfigure}[b]{0.3\textwidth}
            \centering
            \includegraphics[width=\linewidth]{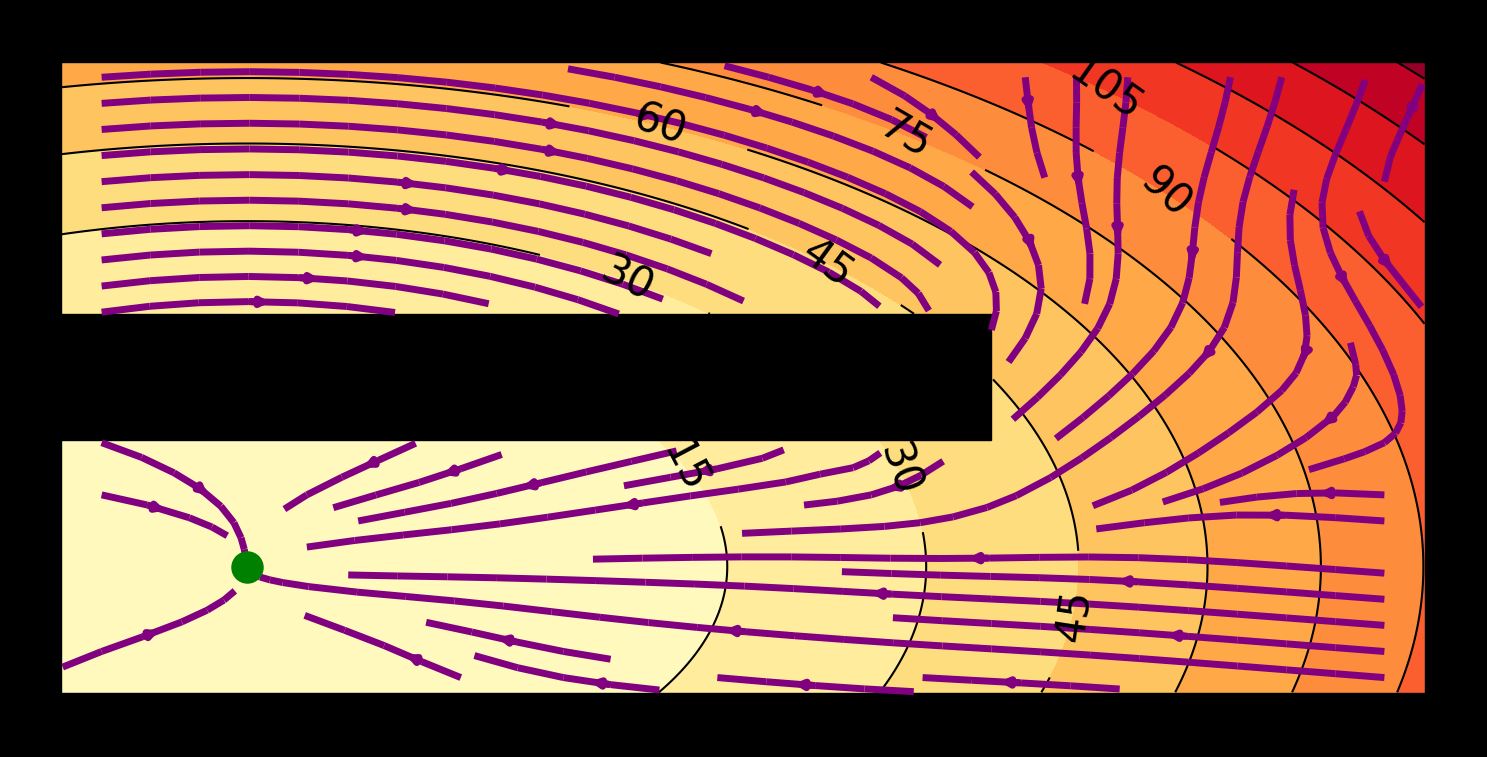}
            \caption{Values and vector field from SEDS with obstacle avoidance.}
            \label{fig:Hall_SEDS}
        \end{subfigure}
        \hfill
        \begin{subfigure}[b]{0.3\textwidth}
            \centering
            \includegraphics[width=\linewidth]{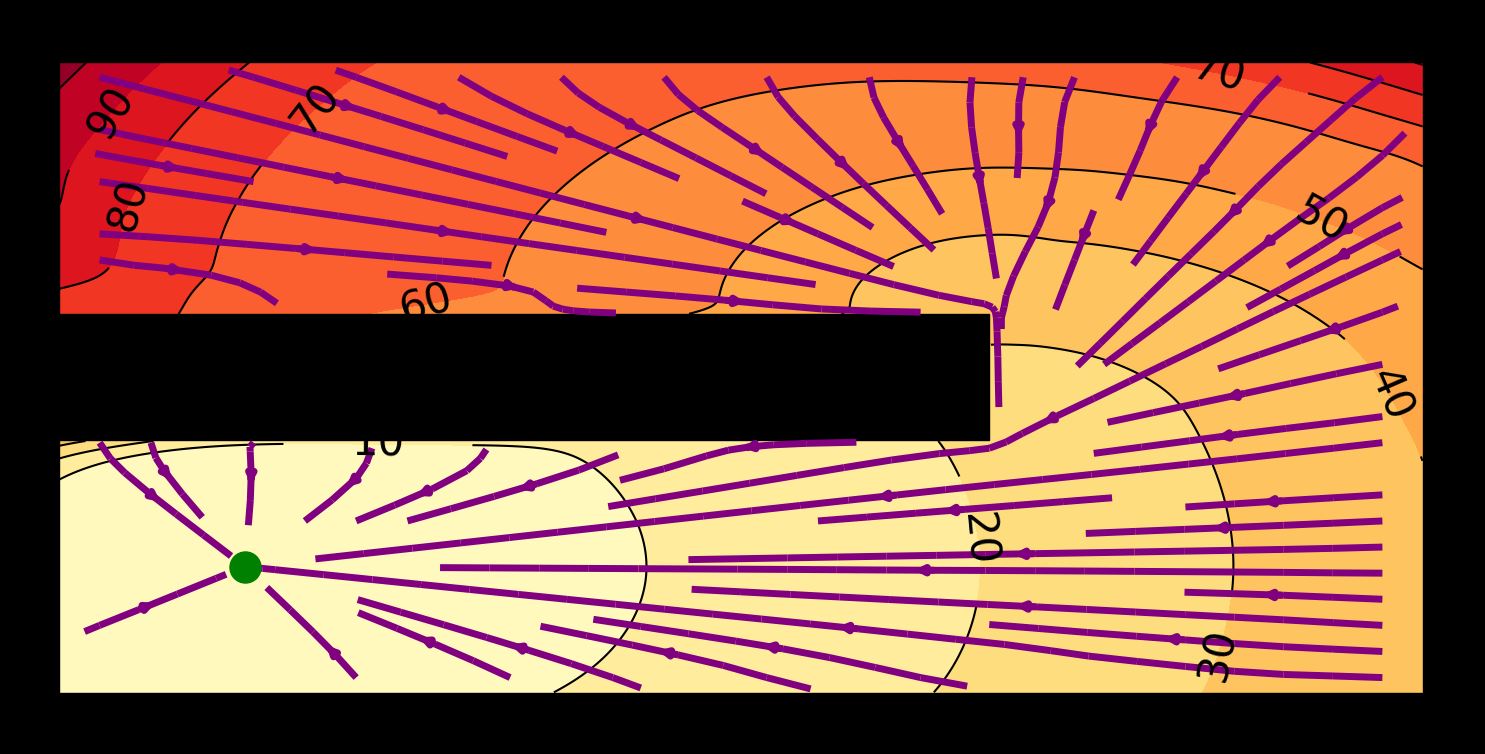}
            \caption{Values and vector field from our method with obstacle avoidance.}
            \label{fig:Hall_Method}
        \end{subfigure}
        \caption{2D hallway environment. The red arrows denote the velocities. The background colour gradient and contour lines show the Lyapunov values for the given position, and the purple streamlines show the motion field of the policy.}
        \label{fig:2d_hallway}
  \end{figure*}
  \begin{figure*}[t]   
        \begin{subfigure}[b]{0.3\textwidth}
            \centering
            \includegraphics[width=\linewidth]{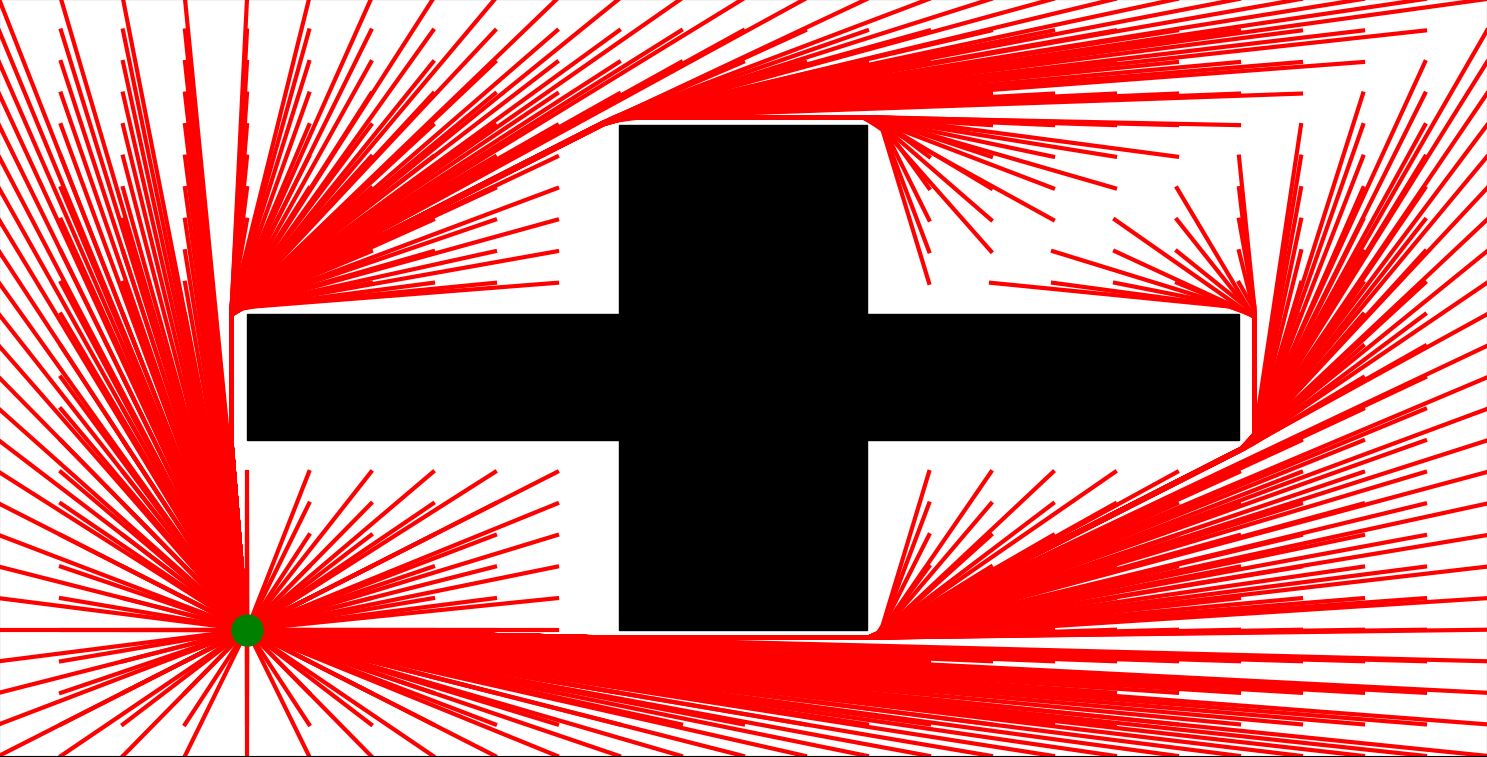}
            \caption{Demonstrations (red arrows) with the goal is marked with a green dot.}
            \label{fig:Cross_Demo}
        \end{subfigure}
        \hfill
        \begin{subfigure}[b]{0.3\textwidth}
            \centering
            \includegraphics[width=\linewidth]{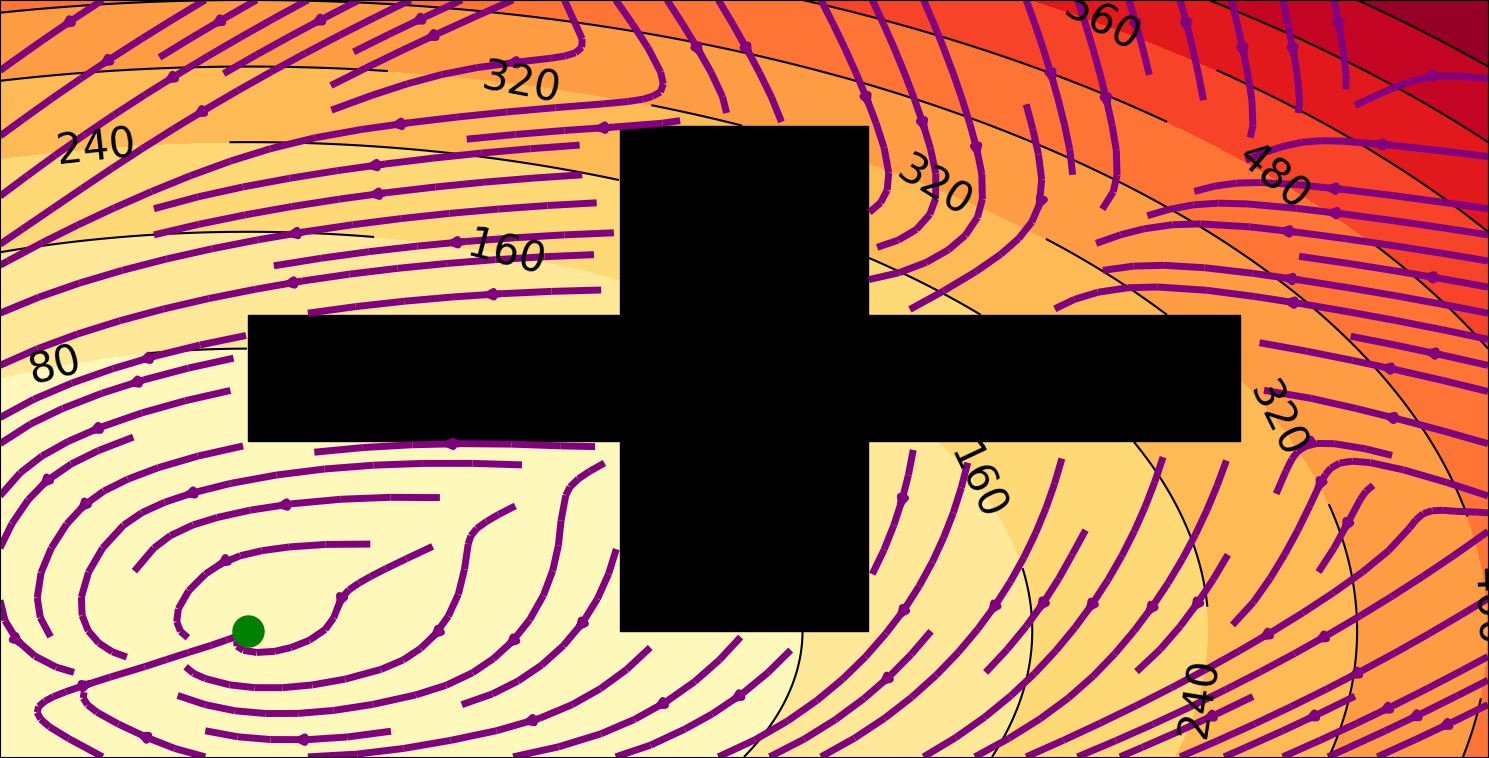}
            \caption{Values and vector field from SEDS with obstacle avoidance}
            \label{fig:Cross_SEDS}
        \end{subfigure}
        \hfill
        \begin{subfigure}[b]{0.3\textwidth}
            \centering
            \includegraphics[width=\linewidth]{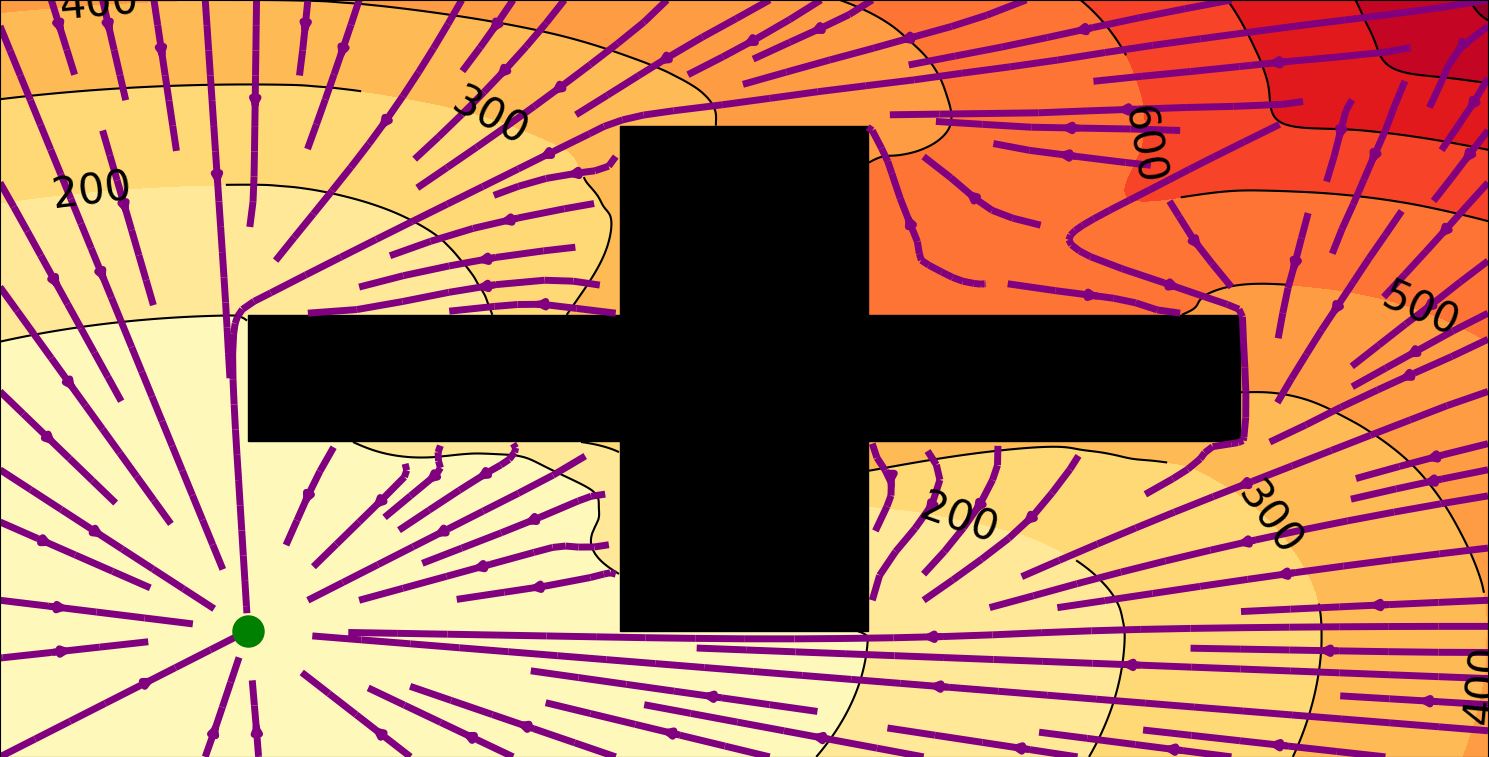}
            \caption{Values and Vector field from our Method with obstacle avoidance}
            \label{fig:Cross_Method}
        \end{subfigure}
        \caption{2D cross environment. The red arrows denote the velocities. The background colour gradient and contour lines show the Lyapunov values for the given position, and the purple streamlines show the motion field of the policy.}
        \label{fig:2d_cross}
    \end{figure*}
    
        \noindent 
    As a proof-of-concept, the method was first validated on 2D examples.
    We designed two scenarios: (a) a hallway case and (b) a cross-shaped case, which can be viewed in Fig.~\ref{fig:2d_hallway} and in Fig.~\ref{fig:2d_cross}. 
    The target position is marked as the green dot.
    
    We used the automatic data collection method in Section III-A to collect 75 and 2818 demonstrations for the hallway and cross environments respectively, where the initial position is a valid position from a grid on the environment around the target position. The collected demonstrations can be viewed as the red arrows in Fig.~\ref{fig:Hall_Demo} and in Fig.~\ref{fig:Cross_Demo}. 
    

    The baseline uses a quadratic Lyapunov function. 
    Fig.\ref{fig:Hall_SEDS} and Fig.\ref{fig:Cross_SEDS} are the results for the baseline, where the background color is the Lyapunov function values and the purple streamlines are the vector fields. 
    In Fig.~\ref{fig:Hall_SEDS}, 
    the Lyapunov function cannot explain the demonstrations at the top left corner.
    Similarly, in  Fig.\ref{fig:Cross_SEDS}, the top right quadrant fails due to the trajectory moving away from the target (which violates the quadratic Lyapunov conditions).

    For both cases, our proposed method finds a solution that brings the system to the target position no matter the starting position, as seen in Fig.~\ref{fig:Hall_Method} and Fig.~\ref{fig:Cross_Method}. 
    In terms of the Lyapunov conditions, we see clearly that the target position has a value of $0$, the values are all above $0$ for the environment and the policy motion field is in the direction of decreasing values, thus satisfying the criteria for stability. 
    
\begin{figure}[t]   
        \centering
        \includegraphics[width=0.9\linewidth]{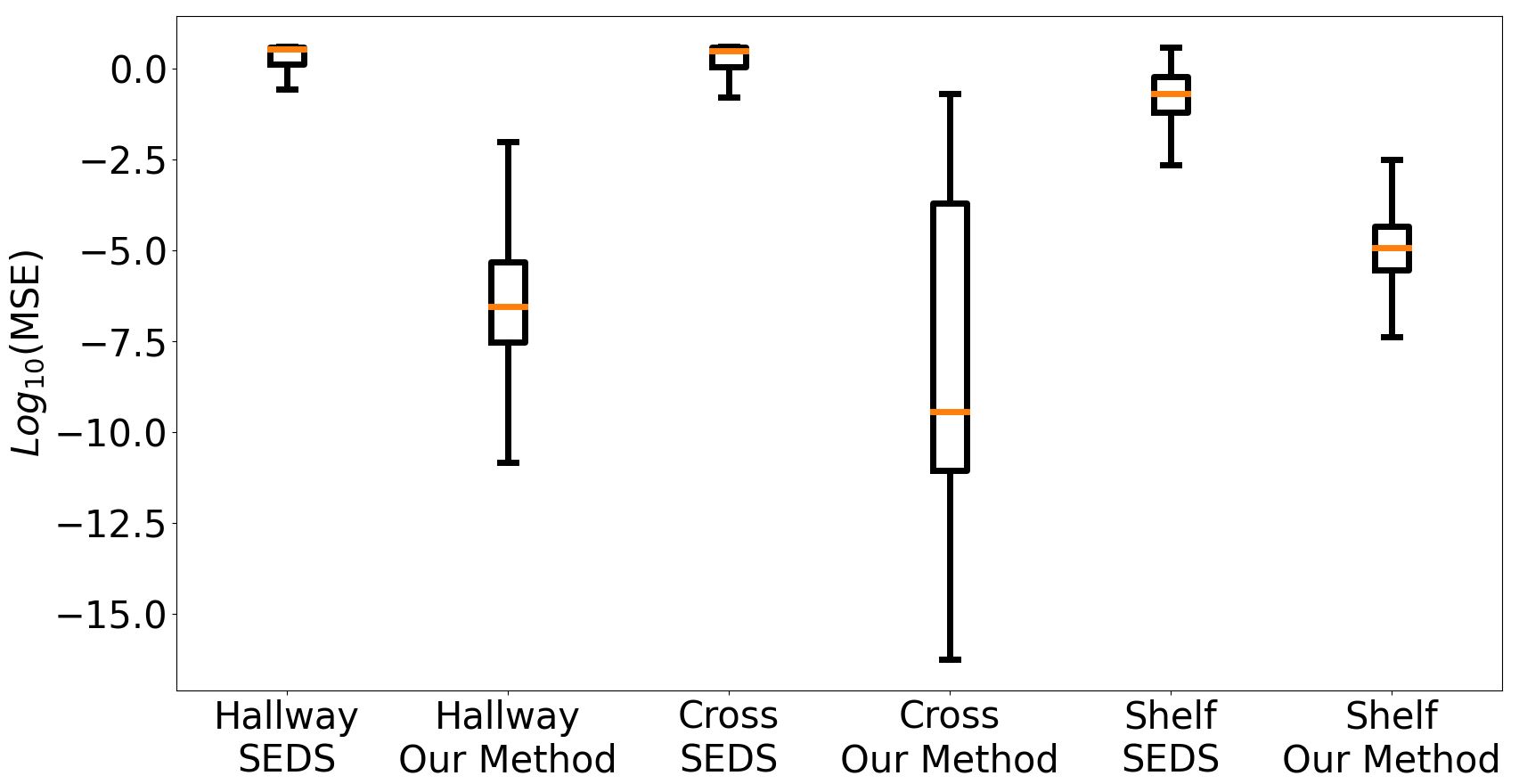}
        \caption{Prediction error of the learned policy using the baseline and our method on the validation set. The y-axis is the \gls{MSE} in log scale.}    
        \label{fig:boxplot}
    \end{figure}

    As for the prediction error, we deploy the policy on our validation demonstration dataset and compare the prediction velocities with the demonstrations. 
    The errors for our method and the baseline on the 2D environments are summarized in Fig~\ref{fig:boxplot}. 
    We can see that our method has a much lower MSE than the baseline. 
    This is due to the policy not being restricted by the quadratic Lyapunov function.

\subsection{Experiment 2: Manipulation in Simulation}
\begin{figure}
        \centering
        \begin{subfigure}[b]{0.51\linewidth}
            \centering
            \includegraphics[width=\linewidth]{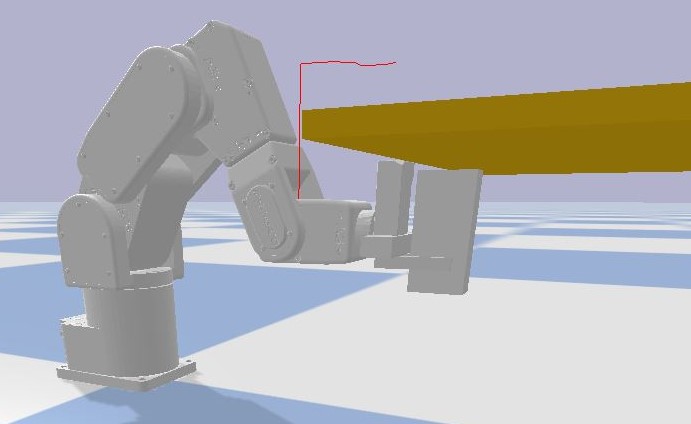}
            \caption{Pybullet simulation}
            \label{fig:MecaBullet}
        \end{subfigure}
        \hfill
        \begin{subfigure}[b]{0.475\linewidth}
            \centering
            \includegraphics[width=0.9\linewidth]{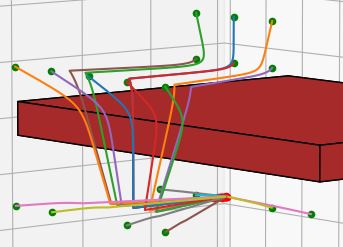}
            \caption{Demonstrations}
            \label{fig:3D_Demos}
        \end{subfigure}
        \caption{Simulation and data for the shelf experiment. (a) Mecademic Meca500 in Pybullet simulation (b) Example data collected from trajectory optimization with multiple initial positions (green) and a target position (orange). }
        \label{fig:3D_experiment}
    \end{figure}

    \noindent
    Next, we explore the performance of our method in a manipulation task where the demonstrations are defined in task space. 
    We use Pybullet~\cite{Pybullet} to simulate the environment and use the Mecademic Meca500 as the robot to perform the task.
    The task is to take an object from the top of a shelf to a target position under the shelf  (see Fig.~\ref{fig:MecaBullet}). 
    
    We collected $1209$ demonstrations by sampling a grid of start positions and select valid positions for trajectory optimization. Some example trajectories are shown in Fig.~\ref{fig:3D_Demos}.
    The network architecture used to learn the model is found in Appendix. 
    The obstacle avoidance module was added so that the payload does not collide with the shelf.

\begin{figure}[t]   
        \centering
        \hfill
        \begin{subfigure}[b]{0.475\linewidth}
            \centering
            \includegraphics[width=0.9\linewidth]{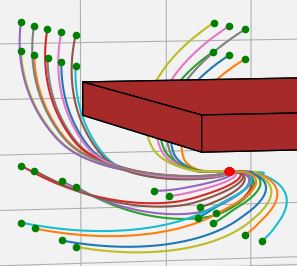}
            \caption{Baseline without collision avoidance}
            \label{fig:3D_SEDS_Rollout_NoObs}
        \end{subfigure}
        \hfill
        \begin{subfigure}[b]{0.475\linewidth}
            \centering
            \includegraphics[width=0.9\linewidth]{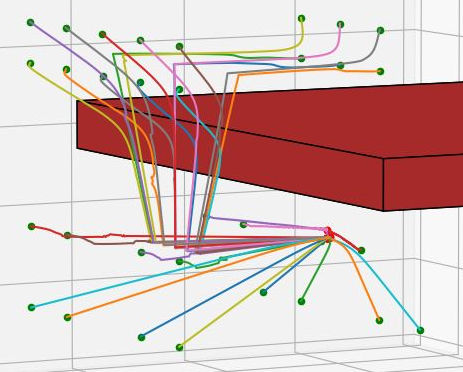}
            \caption{Our method without collision avoidance}
            \label{fig:3D_Method_Rollout_NoObs}
        \end{subfigure}
        \hfill
        \begin{subfigure}[b]{0.475\linewidth}
            \centering
            \includegraphics[width=0.9\linewidth]{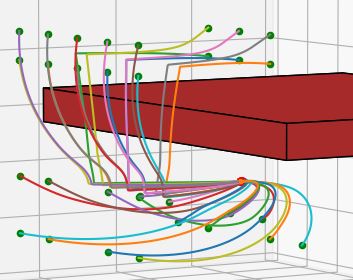}
            \caption{Baseline  with collision avoidance}
            \label{fig:3D_SEDS_Rollout}
        \end{subfigure}
        \hfill
        \begin{subfigure}[b]{0.475\linewidth}
            \centering
            \includegraphics[width=0.9\linewidth]{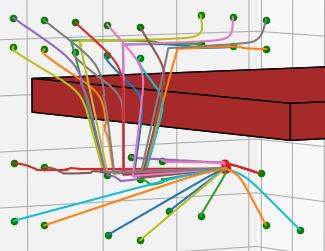}
            \caption{Our method with collision avoidance}
            \label{fig:3D_Method_Rollout}
        \end{subfigure}
     \caption{Planned paths by the learned Policy with \gls{SEDS} and our method where the green points show the initial positions and the orange point shows the goal position. }
     \label{fig:3D_Policy_Rollouts}
    \end{figure}

    The results are summarized in Fig.~\ref{fig:3D_Policy_Rollouts}.
    The outcome of the baseline does not match the demonstrations and collide with the shelf (Fig.~\ref{fig:3D_SEDS_Rollout_NoObs}).
    With collision avoidance, the motion looks closer to the demonstrations, but solely due to the modulation forcing the movement in the same direction of the obstacle (Fig.~\ref{fig:3D_SEDS_Rollout}).
     From Fig.~\ref{fig:3D_Method_Rollout_NoObs} and Fig.~\ref{fig:3D_Method_Rollout}, our method reproduces the motion with or without the obstacle avoidance module. 
    In all cases, the policy rollouts reach the goal.
    From the prediction error in Fig.~\ref{fig:boxplot}, we see that, again, our method has much lower MSE than the baseline.
    
\begin{figure}[t]   
        \centering
        \begin{subfigure}[b]{0.48\linewidth}
            \centering
            \includegraphics[width=\linewidth]{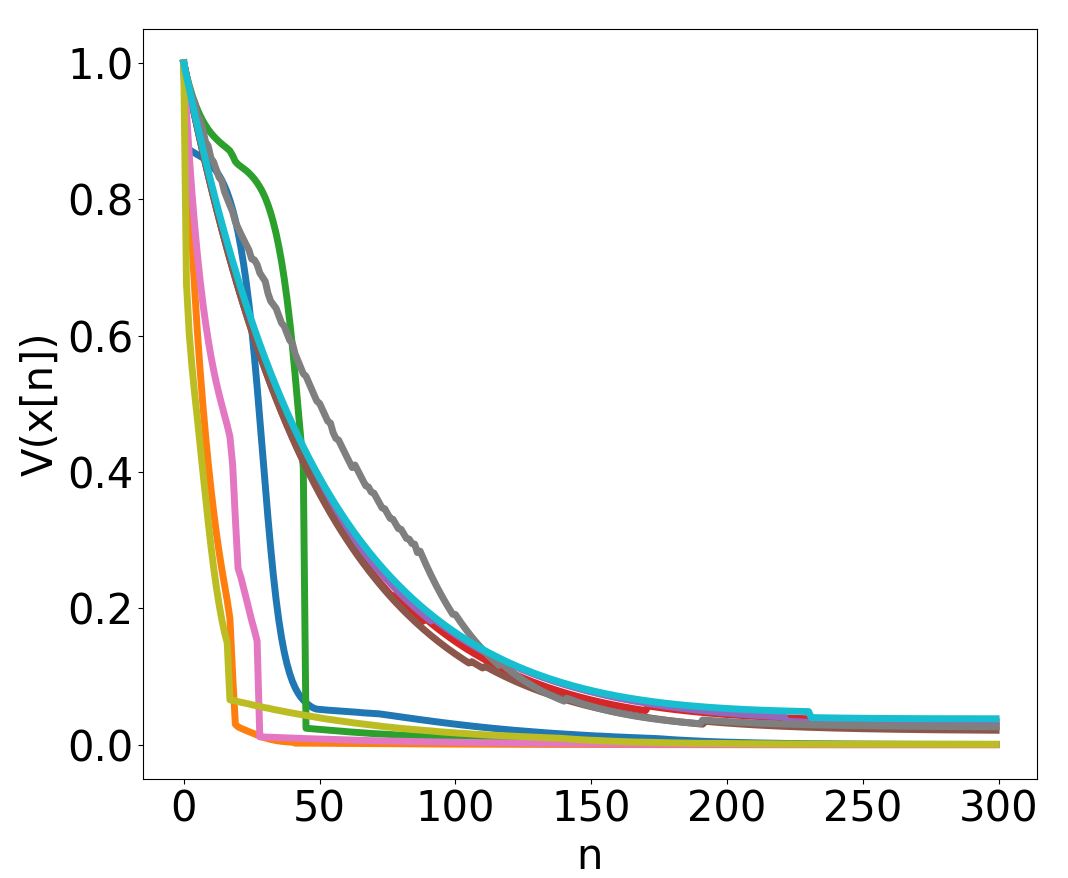}
            \caption{Demonstrations}
            \label{fig:3D_Demo_V}
        \end{subfigure}
        \hfill
        \begin{subfigure}[b]{0.48\linewidth}
            \centering
            \includegraphics[width=\linewidth]{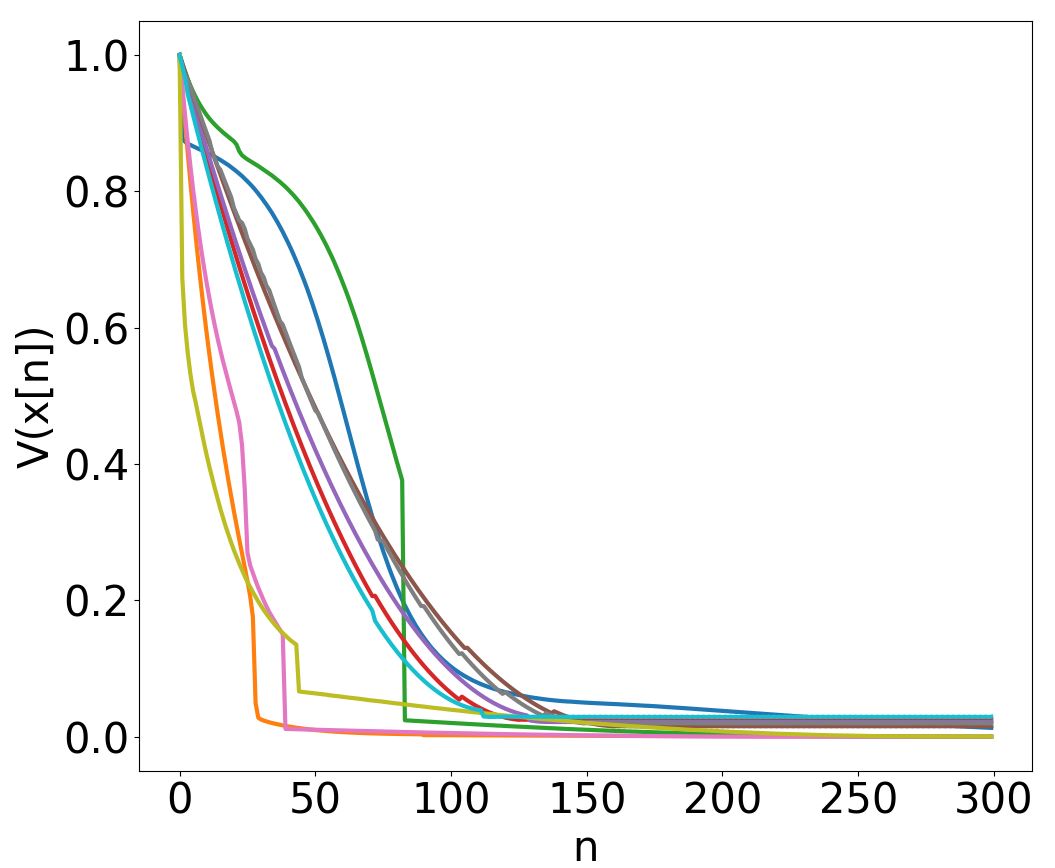}
            \caption{Policy Rollout}
            \label{fig:3D_Policy_V}
        \end{subfigure}
        \caption{Trajectories of the learned Lyapunov function outputs $\lyapFunc{\pos{.}}$ normalized by the output at the initial positions $\lyapFunc{\pos{0}}$, where x-axis is the time-step $n$ and y-axis is the Lyapunov function output $\lyapFunc{\pos{n}}$.}
        \label{fig:3D_V}
    \end{figure}

       Fig.~\ref{fig:3D_V} is a visualization of our learnt Lyapunov function.
    The x-axis is the time-step $n$, and the y-axis is the Lyapunov function output $\lyapFunc{\pos{n}}$ normalized by the output at initial positions $\lyapFunc{\pos{0}}$.
    Fig.~\ref{fig:3D_Demo_V} and Fig.~\ref{fig:3D_Policy_V} are the outputs of $\lyapFunc{\pos{n}}$ with the demonstration data and with the policy rollout, respectively. 
    We can see that the Lyapunov function outputs are positive and monotonically decreasing until they reach $0$ at the goal. 
    This shows that our method found a suitable Lyapunov function that explains the data. 
    
    We also look at the effect of perturbations on the policy. 
    We find that the policy can robustly recover from external perturbations on the robot, since the policy learns a state dependant motion field from the demonstrations and is guaranteed to reach the goal  \href{https://youtu.be/_NzE6LxoSx4}{(see the supplementary video https://youtu.be/\_NzE6LxoSx4)}.
    
\subsection{Experiment 3: Manipulation with a real robot}
    \noindent
    We deploy the policy learnt from Experiment~2 on a robotic hardware, the Meca500 robot from Mecademic. 
    Similar to the simulation, the task is to bring an object from anywhere in the shelf environment to a goal position below the shelf. 
  
    We start the robot from several initial positions and deploy the policy  to bring the system to the goal. 
    The policy is used as the motion planner and provides the physical robot the sequence of task-space via points toward the goal. 
    Then, this is controlled through the standard inverse kinematics control.
    Fig.~\ref{fig:MecaEnv} is an example of the outcome; where Fig.~\ref{fig:MecaInit} shows an example of initial position, Fig.~\ref{fig:MecaTraj} is the policy rollout (red), and Fig.~\ref{fig:MecaGoal} shows that the robot reaches the target position.
    
    The performance of the policy on the real-world robot can be viewed in the supplementary video. 
    We see that the robot is able to perform the task without making contact with the shelf. 
    In all cases, the policy performs the same in task-space on the real-world system as it did in simulation. 
    
    Our method can achieve sim-to-real transfer reliably.
    This is due to our policy being kinematically planned, (i.e., mapping the positions to the velocities), which is less prone to modelling error.
    Also, when the Lyapunov stability conditions are enforced, the movement are restricted to the directions of demonstrations and less likely to diverge. 
    

\section{CONCLUSION \& DISCUSSION}
\label{sec:Conclusion}

\noindent
    We propose a novel method for learning a Lyapunov function and a policy using a single neural network through imitation learning.
    Our method is able to learn a policy that satisfy the Lyapunov stability conditions and reproduce the demonstrations. 
    With our extension of the previous collision avoidance module, our method is capable of avoiding collisions between convex representations of the robot and environment obstacles. 
    The policies were successfully applied to simulated environments and a real-world scenario.

    In future work, the obstacle information will be incorporated into the learned neural network. 
    This will remove the need for the obstacle avoidance module to prevent collisions and have the obstacles be learned by the network. 
    With this improvement, the gradient of the Lyapunov function can have information for how to move away from obstacles and prevent colliding with them. 
    This would make our method self-contained and capable of obstacle avoidance on its own. 


\section*{APPENDIX}
\label{appendix:ModelStruct}
    \setcounter{table}{0}
    \makeatletter 
    \renewcommand{\thetable}{A\@arabic\c@table}
    \makeatother
\noindent
For the baseline, the data were automatically clustered into $6$, $11$ and $3$ Gaussian Mixture Model (GMM) for the hallway, cross, and shelf environment respectively.
Thus, is required to learn $6$, $11$ and $3$ linear dynamical systems for each segment.
    
    For our experiments, the activation function used for all the neurons is the tanh function. 
    The architectures used can be found in Table~\ref{table:NNstructures}. 
    \begin{table}[h]
        \caption{The neural network structures used for each experiment. The information is the number of neurons in each layer: input layer, hidden layers, output layer.}
        \centering
        \begin{tabular}{|c|c|c|}
            \hline
            Experiment & Neural Network Architecture \\ \hline \hline
            Hallway & $\left( 2, 128, 128, 128, 1 \right)$ \\ \hline
            Cross & $\left( 2, 128, 128, 128, 1 \right)$ \\ \hline
            Shelf & $\left( 3, 256, 256, 256, 256, 1 \right)$\\ \hline
        \end{tabular}
        \label{table:NNstructures}
    \end{table}

\newpage
\addcontentsline{toc}{section}{References}
\bibliographystyle{IEEEtran}
\bibliography{references}

\end{document}